\documentclass{article}
\usepackage{spconf,amsmath,graphicx}
\usepackage{multirow}% http://ctan.org/pkg/multirow
\usepackage{hhline}% http://ctan.org/pkg/hhline
\usepackage{algorithm}
\usepackage[noend]{algpseudocode}
\usepackage{hyperref}

\usepackage{float}

\title{Max-pooling Loss Training of Long Short-Term Memory Networks for Small-footprint Keyword Spotting}
\makeatletter
\def\@name{ \emph{Ming Sun$^{1}$, Anirudh Raju$^{2}$, George Tucker$^{3*}$\thanks{*Work conducted while the author was at Amazon.com}, Sankaran Panchapagesan$^{2}$, Gengshen Fu$^{1}$},  \\ \emph{Arindam Mandal$^{2}$, Spyros Matsoukas$^{1}$, Nikko Strom$^{4}$, Shiv Vitaladevuni$^{1}$}}
\makeatother
\address{${}^{1}$Amazon.com, Cambridge, MA, USA\\  ${}^{2}$Amazon.com, Sunnyvale, CA, USA\\ ${}^{3}$Google Brain, Mountain View, CA, USA\\ ${}^{4}$Amazon.com, Seattle, WA, USA\\ \{mingsun,ranirudh,panchi,gengshef,arindamm,matsouka,nikko,shivnaga\}@amazon.com, gjt@google.com}

\begin{document}
\maketitle
\begin{abstract}
We propose a max-pooling based loss function for training Long Short-Term Memory (LSTM) networks for small-footprint keyword spotting (KWS), with low CPU, memory, and latency requirements. The max-pooling loss training can be further guided by initializing with a cross-entropy loss trained network. A posterior smoothing based evaluation approach is employed to measure keyword spotting performance. Our experimental results show that LSTM models trained using cross-entropy loss or max-pooling loss outperform a cross-entropy loss trained baseline feed-forward Deep Neural Network (DNN). In addition, max-pooling loss trained LSTM with randomly initialized network performs better compared to cross-entropy loss trained LSTM. Finally, the max-pooling loss trained LSTM  initialized with a cross-entropy pre-trained network shows the best performance, which yields $67.6\%$ relative reduction compared to baseline feed-forward DNN in Area Under the Curve (AUC) measure.
\end{abstract}


\section{Introduction}
\label{sec:intro}

Keyword spotting has been an active research area for decades. Different approaches have been proposed to detect the words of interest in speech utterances. As one solution, a general large vocabulary continuous speech recognition (LVCSR) system is applied to decode the audio signal, and keyword searching is conducted in the resulting lattices or confusion networks \cite{miller07,parlak08,chen13,tsakalidis14}. These methods require relatively high computational resources for the LVCSR decoding, and also introduce latency.

Small-footprint keyword spotting systems have been increasingly attracting attention. Voice assistant systems such as Alexa on Amazon Echo deploy a keyword spotting system on device, and only stream audio to the cloud for LVCSR when the keyword is detected on device. For such applications, accurate on-device keyword spotting running with low CPU and memory is critical \cite{sun15}. It needs to run with high recall to make devices easy to use, while having  low false accepts to mitigate privacy concerns. Latency has to be low as well. A traditional approach employs Hidden Markov Model (HMM) to model both keyword and background \cite{rose90,wilpon90,wilpon91}. The background includes non-keyword speech, or non-speech noise etc. This background model is also named filler model in some literatures. It could involve loops over simple speech/non-speech phones, or for more complicated cases, normal phone set or confusing word set. Viterbi decoding is used to search the best path in the decoding graph. The keyword spotting decision can be made based on the likelihood comparison of keyword and background models. Gaussian Mixture Model (GMM) was commonly used in the past to model the observed acoustic features. With DNN becoming mainstream for acoustic modeling, this approach can be extended to include discriminative information by incorporating a hybrid DNN-HMM decoding framework \cite{panchapagesan16}.

In recent years, there are keyword spotting systems built on DNN or Convolutional Neural Network (CNN) directly, with no HMM involved in the system \cite{chen14,nakkiran15,sainath15,tucker16}. During decoding time, framewise keyword posteriors are smoothed. The system is triggered when smoothed keyword posteriors exceed a pre-defined threshold. The trade off between balancing false rejects and false accepts can be performed by tuning the threshold. Context information is taken care of by stacking frames as input. Some keyword spotting systems are built on Recurrent Neural Network (RNN) directly. Particularly, bidirectional LSTM is used to search for keywords in audio streams when latency is not a hard constraint \cite{fernndez07,wollmer13,baljekar14,sundar15}.

We are interested in a small-footprint keyword spotting system that runs on low CPU and memory utilization, with low latency. This low latency constraint makes bidirectional LSTM not a proper fit in principle. Instead, we focus on training a unidirection LSTM model using two different loss functions: cross-entropy loss  and max-pooling based loss \cite{scherer10}. Applying the max-pooling loss function to LSTM training for keyword spotting is the main contribution of this paper.

During decoding time, the system is triggered when the keyword posterior smoothed by averaging the output of a sliding window is above a threshold. Considering the practical use case, our keyword spotting system is designed to lock out for some time after each detection, to avoid unnecessary false accepts and reduce decoding computational cost.

The remaining part of this paper is organized as follows: Section \ref{sec:system} describes our LSTM based keyword spotting system, which includes the LSTM model, training loss functions and performance evaluation details. Experimental setup and results are included in Section \ref{sec:exp}. Section \ref{sec:conc} is for conclusion and future work.

\section{System Overview}\label{sec:system}
\begin{figure}[htb]
	\centering
	\includegraphics[width=0.5\textwidth]{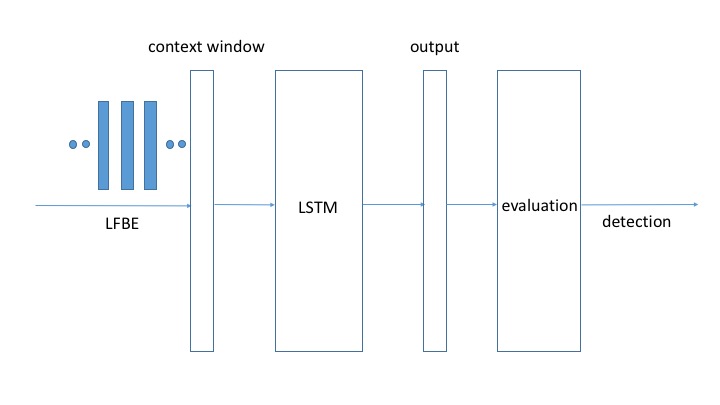}
\caption{Keyword spotting system}
\label{fig:sys}
\end{figure}

As shown in Figure \ref{fig:sys}, Log Mel Filter-Bank Energies (LFBEs) are used as input acoustic features for our keyword spotting system. We extract 20 dimensional LFBEs over 25ms frames with a 10ms frame shift. The LSTM model is used to process input LFBEs. Our system has two targets in the output layer: non-keyword and keyword. The output of the keyword spotting system is passed to an evaluation module for decision making.
\subsection{LSTM}\label{sec:lstm}
\begin{figure}[htb]
	\centering
	\includegraphics[width=0.5\textwidth]{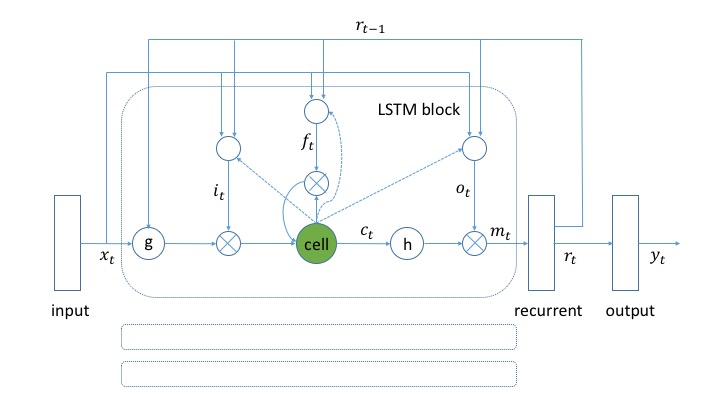}
\caption{Architecture of LSTM with projection layer}
\label{fig:lstm}
\end{figure}

Different from feed-forward DNN networks, RNNs contain cyclic connections which can be used to model sequential data . This makes RNNs a natural fit to model temporal information within continuous speech frames. However, traditional RNN structures suffer from the vanishing gradient problem, which prevents them from effectively modeling long context in the data. To overcome this, LSTMs contain memory blocks \cite{hochreiter97,gers02}. Each block contains one or more memory cells, as well as input, output and forget gates. These three gates control the information flow within the associated memory block. Sometimes a projection layer is added on top of the LSTM output, to reduce model complexity \cite{sak14}.  A typical LSTM component with projection layer is shown in Figure \ref{fig:lstm}. For the sake of clarity, a single LSTM block is  shown here.

Given a sequence of T frames $\mathbf{X} = (\mathbf{x}_1,\ldots,\mathbf{x}_T)$, let ${\mathbf{i},\mathbf{o},\mathbf{f},\mathbf{c}}$ denote the input, output, forget gates, and the memory cell, and ${\mathbf{Y}=(\mathbf{y}_1,\ldots,\mathbf{y}_T)}$ be the output. The LSTM computes the gate activations and output at time $t$ as follows:

\begin{subequations}
\begin{align*}
\mathbf{i}_t = \sigma(\mathbf{W}_{ix}\mathbf{x}_t + \mathbf{W}_{ir}\mathbf{r}_{t-1} + \mathbf{W}_{ic}\mathbf{c}_{t-1} + \mathbf{b}_i) \\
\mathbf{f}_t = \sigma(\mathbf{W}_{fx}\mathbf{x}_t + \mathbf{W}_{fr}\mathbf{r}_{t-1} + \mathbf{W}_{fc}\mathbf{c}_{t-1} + \mathbf{b}_f) \\
\mathbf{c}_t = \mathbf{f}_t \odot \mathbf{c}_{t-1} + \mathbf{i}_t \odot g(\mathbf{W}_{cx}\mathbf{x}_t + \mathbf{W}_{cr}\mathbf{r}_{t-1} + \mathbf{b}_c) \\
\mathbf{o}_t = \sigma(\mathbf{W}_{ox}\mathbf{x}_t + \mathbf{W}_{or}\mathbf{r}_{t-1} + \mathbf{W}_{oc}\mathbf{c}_{t} + \mathbf{b}_o) \\
\mathbf{m}_t = \mathbf{o}_{t} \odot h(\mathbf{c}_t) \\
\mathbf{r}_t = \mathbf{W}_{rm} \mathbf{m}_t \\
\mathbf{y}_t = \phi(\mathbf{W}_{yr} \mathbf{r}_t + \mathbf{b}_y)
\end{align*}
\end{subequations}

Here $\mathbf{W}_{\ast}$ matrices label the connection weights. E.g., $\mathbf{W}_{ix}$, $\mathbf{W}_{ir}$ and $\mathbf{W}_{ic}$ represent the weight matrices from the input $\mathbf{x}$, recurrent feedback $\mathbf{r}$ and cell $\mathbf{c}$ respectively. Note that the peephole connections $\mathbf{W}_{ic}$, $\mathbf{W}_{fc}$ and $\mathbf{W}_{oc}$ are diagonal matrices. The $\mathbf{b}_{\ast}$ terms represent the bias vectors for different components of the model. E.g., $\mathbf{b}_i$ is the bias for input gate activation. 

A projection layer is added to the LSTM output. That is, $\mathbf{W}_{rm}$ linearly maps $\mathbf{m}_t$ to a lower dimensional representation $\mathbf{r}_t$, which is the recurrent signal. The network output $\mathbf{y}_t$ is computed based on the projection layer output $\mathbf{r}_t$ as well.

Regarding the activation functions, we use logistic sigmoid function as ${\sigma()}$ for gate activations, $\tanh$ as $g()$ and $h()$ for cell input and output, and softmax as $\phi()$ for output layer. $\odot$ is the element-wise product of vectors.

Finally, the complexity of the model described above can be calculated as
\begin{equation}
n = n_c \times n_r \times 4 + n_i \times n_c \times 4 + n_r \times n_o + n_c \times n_r + n_c \times 3
\end{equation}
where $n_c$ is the number of memory cells (we only consider the case of single memory cell per block, thus here $n_c$ is also the number of memory blocks), $n_r$ is the dimension of projection layer, ${n_i}$ and ${n_o}$ denote the dimension of input and output respectively.

\subsection{Loss functions}\label{sec:loss}
For our experiments, we consider two different types of loss functions: cross-entropy loss and max-pooling loss.

\subsubsection{Cross-entropy}\label{sec:xent}
Cross-entropy (xent) has been widely applied as a loss function for DNN and RNN training \cite{bishop06}. Let $K$ be the total number of classes. Given a sequence of $T$ frames $\mathbf{X} = (\mathbf{x}_1,\ldots,\mathbf{x}_T)$, where $\mathbf{x}_t$  is the feature vector of the $t$th frame, let $\mathbf{y}_t = (y_t^1,\ldots,y_t^K)$ denote the $K$-dimensional output of the network for $\mathbf{x}_t$, and let $\mathbf{z}_t = (z_t^1,\ldots,z_t^K)$ denote the corresponding target vector. The cross-entropy loss for the $t$th frame is calculated as follows:
\begin{equation}
\mathcal{L}_{t}^{xent} = - \sum_{k=1}^{K} z_t^k \ln y_t^k
\end{equation}
The $1$-of-$K$ coding is usually used for target vector $\mathbf{z}_t$. That is, if the $t$th frame vector $\mathbf{x}_t$ is aligned with class $k$, the $K$-dimensional vector $\mathbf{z}_t$ has value $1$ for the $k$th element, with all other elements  being $0$. Let $k^{\dagger}_t$ denote the aligned class for the $t$th frame. The cross-entropy loss for the $t$th frame can be formulated as
\begin{equation}
\mathcal{L}_{t}^{xent} = -\ln y_t^{k^{\dagger}_t}
\end{equation}

Then the cross-entropy loss for the whole $T$ frame sequence is:
\begin{equation}
\mathcal{L}_{T}^{xent} = \sum_{t=1}^{T} \mathcal{L}_{t}^{xent} = -\sum_{t=1}^{T} \ln y_t^{k^{\dagger}_t}
\end{equation}

\subsubsection{Max-pooling}\label{sec:maxpool}
We propose to train the LSTM for keyword spotting using a max-pooling based loss function. Given that the LSTM has the ability to model long context information, we hypothesize that there is no need to teach the LSTM to fire every frame within the keyword segment. Instead, we want to teach the LSTM to fire at its highest confidence time. The LSTM should fire near the end of keyword segment in general, where it has seen enough context to make a decision. A simple way is to back-propagate loss only from the last frame or last several frames for updating the weights. But our initial experiments indicate that the LSTM does not learn much from this scheme. Hence we employ a max-pooling  based loss function to let the LSTM pick the most informative keyword frames to teach itself. This also helps mitigate issues potentially caused by inaccurate frame alignment around keyword segment boundaries. Max-pooling loss can be viewed as a transition from frame-level loss to segment-level loss for keyword spotting model training. 

Alternative segment-level loss functions include different statistics of frame-level keyword posteriors within a keyword segment, e.g., the geometric mean etc. There have been literatures on training LSTMs using Connectionist Temporal Classification (CTC) \cite{fernndez07,wollmer13,baljekar14,graves05} for keyword spotting tasks as well. In addition, architectures that combine LSTMs and CNNs have been applied to different tasks \cite{ng15, xu15}. Typically LSTM is added on top of CNN layers, where CNN layers with pooling are used to extracted features as LSTM input, and LSTM output is used for prediction.

Let $Q$ denote the cardinality of target keyword set. When we consider word level labels, there are in total $Q+1$ classes ($K=Q+1$), with one additional class used to label those frames aligned with background. For the $T$ input frames $\mathbf{X} = (\mathbf{x}_1,\ldots,\mathbf{x}_T)$, if there are $P$ keywords instances inside, we use $\mathbf{l}_p$ to denote a continuous frame index range whose frames are aligned with the $p$th keyword. As a result, the $K$-dimensional target vector $\mathbf{z}_p$ is the same for all frames within $\mathbf{l}_p$. 

Let $\mathbf{L} = (\mathbf{l}_1,\ldots,\mathbf{l}_P)$ represent a collection of frame index ranges for all $P$ keywords instances in $\mathbf{X}$, and $\hat{\mathbf{L}}$ be a collection of all the indices for the remaining frames which are not aligned to any keyword (i.e. background frame indices). We use $k^{\dagger}_p$ to represent the target label for frames inside $\mathbf{l}_p$, and $l^{\dagger}_p$ to label the specific one frame within $\mathbf{l}_p$ whose posterior for $k^{\dagger}_p$ is the maximum. The max-pooling loss proposed for the input sequence $\mathbf{X}$ can be calculated as
\begin{equation}
\mathcal{L}_{T}^{maxpool} = - \sum_{t\in \hat{\mathbf{L}}} \ln y_t^{k^{\dagger}_t} - \sum_{p=1}^{P} \ln y_{l^{\dagger}_p}^{k^{\dagger}_p}
\end{equation}
The first item states that we calculate the cross-entropy loss for input frames not aligned to any keyword. The second item shows how we do max-pooling for keyword aligned frames. In more details, for the frames of the $p$th segment (index range $\mathbf{l}_p$), they are aligned to keyword $k^{\dagger}_p$. We only back propagate for a single frame (index $l^{\dagger}_p$) whose posterior for target $k^{\dagger}_p$ is the largest among all frames within current segment $\mathbf{l}_p$, and discard all other frames within current segment. 

\begin{figure*}[htb]
	\centering
	\includegraphics[width=0.7\textwidth]{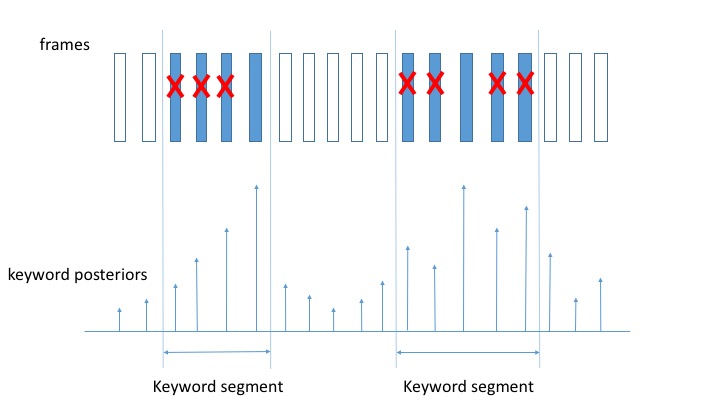}
\caption{Idea of max-pooling loss}
\label{fig:maxpool}
\end{figure*}

The idea of max-pooling loss is shown in Figure \ref{fig:maxpool}, where filled frames are aligned with the keywords, and empty frames are for background. Given an input sequence of frames, within each keyword segment, only the frame which has the maximum posterior for corresponding keyword target is kept, while all other frames within the same keyword segment are discarded. All background frames are kept.

We consider two cases for max-pooling loss based LSTM training: one starts with a randomly initialized model, and the other uses a cross-entropy loss pre-trained model. With a randomly initialized model, max-pooling loss based LSTM training may not learn well in the first few epochs with rather random keyword firing. The idea is to take the advantages of both cross-entropy and max-pooling loss training. With a cross-entropy trained LSTM as the initial model to start max-pooling training, it already learns some basic knowledge about target keywords. This could provide a better initialization point, and faster convergence to a better local optimum.

\subsection{Evaluation Method}\label{sec:eval}
We consider a posteriors smoothing based evaluation scheme. To detect the keyword, given input audio, the system computes smoothed posteriors based on a sliding context window containing $N_{ctx}$ frames. When the smoothed posterior for the keyword exceeds a pre-defined threshold, this is considered as a firing spike. The system is designed to shut down for the following $N_{lck}$ frames. This lockout period of length $N_{lck}$ is for the purpose of reducing unnecessarily duplicated detections during the same keyword segment, as well as reducing decoding computational cost.

For our use case, we allow a short latency period with ${N_{lat}}$ frames after each keyword segment. That is, if the system fires within the ${N_{lat}}$-frame window right after a keyword segment, we still consider the firing as being aligned with the corresponding keyword. This latency window does not introduce significant delay in perception, and it could mitigate the possible issues of inaccurate keyword alignment boundaries in evaluation.

Finally, the first firing spike within each keyword segment plus latency window is considered as a valid detection. Any other firing spikes within the same keyword segment plus latency window, or outside any keyword segment plus latency window, are counted as false accepts. Two metrics are used to measure the system performance: miss rate, which is one minus recall, and false accept rate, which is a normalized value of false accepts.

\begin{figure*}[htb]
	\centering
	\includegraphics[width=0.7\textwidth]{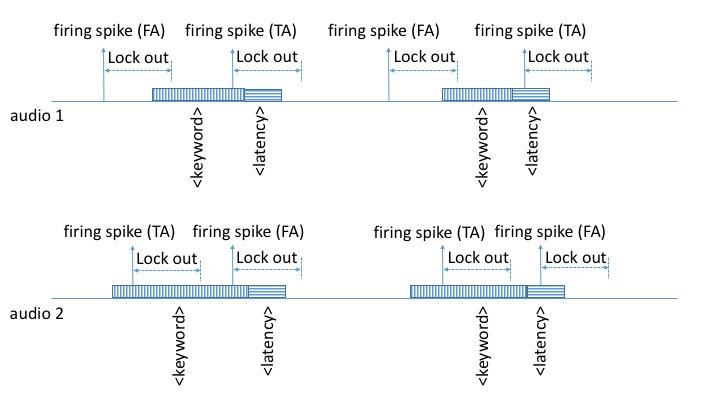}
\caption{Illustration of evaluation method}
\label{fig:eval}
\end{figure*}

Figure \ref{fig:eval} illustrates the idea of our evaluation approach. As examples, there are two input audio streams. The keyword segment length varies depending on the way the keyword is spoken. Each keyword segment is followed by a fixed length latency window. The keyword segments are labeled by blocks with vertical line fill, while the follow-on latency windows are labeled by blocks with horizontal line fill. There is a system lock out period by design after each firing spike. For the first audio, there are two false accepts (FAs) with system firing in the region outside any keyword segment plus latency window. The true accepts (TAs) happen as the first detection in each keyword segment plus latency window. True accepts could happen either in the keyword segment, or in the following latency window. For the second audio, false accepts happen as additional firing spikes within the same keyword segment plus latency window which already has a true accept.

For our system, we use $30$ frames ($N_{ctx}=30$) for posteriors smoothing, $40$ frames ($N_{lck}=40$) as the lockout period, and $20$ frames ($N_{lat}=20$) as the allowed latency window length after each aligned wake word segment. 

\section{Experimental Results}\label{sec:exp}
For our experiments, the word 'Alexa' is chosen as the keyword. We use an in-house far-field corpus which contains far-field data collected under different conditions. This dataset contains an order of magnitude more instances of keyword and background speech utterances than the largest previous studies \cite{chen14,sainath15} for both training and testing. Our data is collected in a far-field environment, which is a more challenging task by nature. Considering the large size of our corpus, the development set partition is sufficient to tune parameters, and the test set partition is large enough to show strong statistical difference. 

Since we only target for one keyword 'Alexa', a binary target set is used for our experiments. Frames of background have target label $0$, while frames aligned with keyword have target $1$. We train a feed-forward DNN model as the baseline based on the model structure and training described in \cite{chen14}, with some adaptations to our experimental setup and use case. We compare it with the LSTM models trained with cross-entropy loss and max-pooling loss.

\subsection{Model Training} \label{sec:model_train}
The GPU-based distributed trainer described in \cite{strom15} is used for our experiments. A performance based learning rate schedule is used for our model training. To elaborate, for each training epoch, if the loss on the dev set degrades compared to the previous epoch, the learning rate is halved, and current epoch is repeated with reduced learning rate. Training process terminates when either the minimal learning rate (for our case, a factor of $0.5^8$ of initial learning rate), or the maximum number of epochs is reached (we limit our training to be $20$ epochs). The initial learning rate and batch size are tuned on the development set.  

The baseline feed-forward DNN has four hidden layers, with 128 nodes per hidden layer. Sigmoid function is used as activation. A stack of 20 frames on the left and 10 frames on the right are used to form an input feature vector. Note that the right context cannot be too large, since it introduces latency. There are in total $\sim129K$ parameters with the DNN model. Layerwise pre-training is used for the DNN. Initial learning rate for DNN training is $0.0005$, and batch size is $256$.

For LSTM training with different loss functions, we use a single layer of unidirectional LSTM with 64 memory blocks and a projection layer of dimension 32. This serves the purpose of low CPU and memory, as well as low latency. For input context, we consider $10$ frames on the left and $10$ frames on the right. Note that we still use $10$ frames as left context for LSTM input, though the LSTM learns past frames' information by definition. By doing this our DNN and LSTM training setup are aligned better for comparison, and past information is further imposed for LSTM training. Our LSTM has $\sim118K$ parameters. For random initialization, the LSTM parameters are initialized with a uniform distribution $U[-0.2, 0.2]$ for weights, and constant $0.1$ for bias. The initial learning rates are chosen to be $0.00001$, $0.00005$ and $0.00005$ for the cases of cross-entropy loss, max-pooling loss with randomly initialized model, and max-pooling loss initialized with a cross-entropy pre-trained model. 

\subsection{System Performance}
\begin{figure}[htb]
	\centering
	\includegraphics[width=0.5\textwidth]{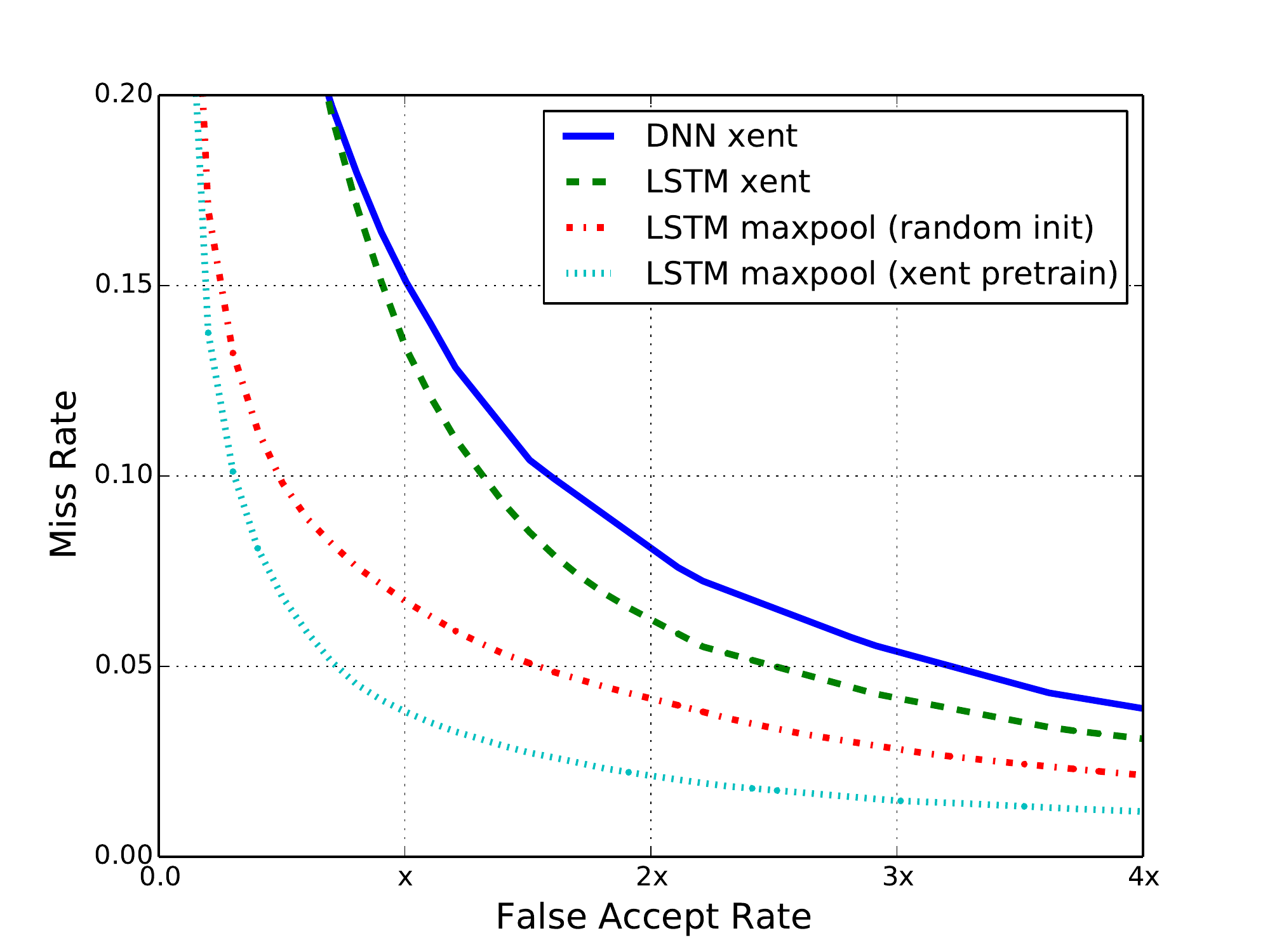}
 
\caption{Performance of DNN and LSTM models}
\label{fig:perf}
\end{figure}

We use the evaluation approach described in Section \ref{sec:eval} on our test dataset. The performance of the DNN and LSTM models are shown in Figure \ref{fig:perf}.

We plot detection (DET) curves in a low miss rate range, i.e., $\leq20\%$ for this case. Here the false accept rate is computed by normalizing the false accept counts with the total number of test data utterances. The x-axis labels false accept rate, and the y-axis labels miss rate. Lower numbers indicate better performance. The blue solid curve represents the baseline feed-forward DNN trained using cross-entropy loss. The LSTM models trained using cross-entropy loss, max-pooling loss with random initialization, and max-pooling loss with cross-entropy pre-training, are labeled by the green dashed, red dash-dot and cyan dotted curves respectively. Absolute numbers of false accepts have been obscured in this paper due to confidentiality reasons. Instead, we plot false accept rates up to a multiplicative constant. The false accept range considered in our experiments is aligned with a low value range which can be considered for production deployment purpose.

In the selected low miss rate range, LSTM models outperform the baseline feed-forward DNN. In terms of different loss functions for LSTM training, max-pooling loss with random initialization is superior to cross-entropy loss. LSTM trained using max-pooling loss with cross-entropy loss pre-training yields the best results. We compute the Area Under the Curve (AUC) numbers for quantitative comparison of different models. AUC is computed on DET curves and hence lower is better. The relative changes of AUC for LSTM models compared to the baseline DNN are summarized in Table \ref{tbl:auc}. Our experimental results indicate that in the $\leq20\%$ low miss rate range, compared to a cross-entropy loss trained baseline DNN, cross-entropy loss trained LSTM results in $34.4\%$ relative reduction in AUC. The LSTM model trained using max-pooling loss with random initialization further shows $48.2\%$ relative reduction in AUC. The best performance comes from the LSTM trained using max-pooling loss with cross-entropy pre-training, which yields $67.6\%$ AUC reduction compared to the baseline DNN. 
\begin{table}[!htbp]
	\centering
	\begin{tabular}{|c|c|c|c|c|}
		\hline
		model & DNN & \multicolumn{3}{c|}{LSTM} \\
		\hline
		\multirow{2}{*}{loss function} & \multirow{2}{*}{xent} & \multirow{2}{*}{xent} & \multicolumn{2}{c|}{maxpool}  \\
		\cline{4-5}
		& & & random init & xent pretrain \\
		\hline
		AUC change & $0\%$ & $-34.4\%$ & $-48.2\%$ & $-67.6\%$ \\
		\hline
	\end{tabular}
	\caption{Relative change of AUC for LSTM models compared to the baseline feed-forward DNN. Lower AUC indicates better performance.}\label{tbl:auc}
\end{table}

\section{Conclusion and Future Work}\label{sec:conc}
We present our work of training a small-footprint LSTM to spot the keyword 'Alexa' in far-field conditions. Two loss functions are employed for LSTM training: one is cross-entropy loss, and the other is max-pooling loss proposed in this paper. A smoothed posterior thresholding approach is used for evaluation. Keyword spotting performance is measured using miss rate and false accept rate. We show that LSTM performs better than DNN in general. The best LSTM system, which is trained using max-pooling loss with cross-entropy loss pre-training, reduces the AUC number by $67.6\%$ in the low miss rate range.

For future work, we plan to add weighting to max-pooling loss based LSTM training, i.e., scale the back-propagated loss for the selected keyword frames. It is of interest to see if LSTM performance can be further improved by varying model structures, e.g., adding additional feed-forward layers on top of the LSTM component. We also plan to benchmark max-pooling loss performance against other segmental level loss functions, e.g., geometric mean of framewise keyword posteriors within each keyword segment, CTC etc, for our keyword spotting experiments. 

\bibliographystyle{IEEEbib}
\bibliography{strings,refs}

\begin{thebibliography}{10}
\bibitem[1]{miller07} Miller, D.R., Kleber, M., Kao, C.L., Kimball, O., Colthurst T., Lowe, S.A., Schwartz, R.M., and Gish, H., ``Rapid and accurate spoken term detection'', in \textit{Proceedings of Annual Conference of the International Speech Communication Association (Interspeech)}, 2007.
\bibitem[2]{parlak08} Parlak, S. and Saraclar, M., ``Spoken term detection for Turkish broadcast news'', in \textit{IEEE International Conference on Acoustics, Speech and Signal Processing (ICASSP)}, pp. 5244-5247, 2008.
\bibitem[3]{chen13} Chen, G., Yilmaz, O., Trmal, J., Povey, D. and Khudanpur, S., ``Using proxies for OOV keywords in the keyword search task'', in \textit{IEEE Workshop on Automatic Speech Recognition and Understanding (ASRU)}, pp. 416-421, 2013.
\bibitem[4]{tsakalidis14} Tsakalidis, S., Hsiao, R., Karakos, D., Ng, T., Ranjan, S., Saikumar, G., Zhang, L., Nguyen, L., Schwartz, R. and Makhoul, J., ``The 2013 BBN vietnamese telephone speech keyword spotting system'', in \textit{IEEE International Conference on Acoustics, Speech and Signal Processing (ICASSP)}, pp. 7829-7833, 2014.
\bibitem[5]{sun15} Sun, M., Nagaraja, V., Hoffmeister, B. and Vitaladevuni, S., ``Model Shrinking for Embedded Keyword Spotting'', in \textit{IEEE 14th International Conference on Machine Learning and Applications (ICMLA)}, 2015.
\bibitem[6]{rose90} Rose, R.C. and Paul, D.B., ``A hidden Markov model based keyword recognition system'', in \textit{IEEE International Conference on Acoustics, Speech, and Signal Processing (ICASSP)}, pp. 129-132, 1990.
\bibitem[7]{wilpon90} Wilpon, J.G., Rabiner, L., Lee, C.H. and Goldman, E.R., ``Automatic recognition of keywords in unconstrained speech using hidden Markov models'', \textit{IEEE Transactions on Acoustics, Speech and Signal Processing}, 38(11):1870-1878, 1990.
\bibitem[8]{wilpon91} Wilpon, J.G., Miller, L.G. and Modi, P., ``Improvements and applications for key word recognition using hidden Markov modeling techniques'', in \textit{IEEE International Conference on Acoustics, Speech, and Signal Processing (ICASSP)}, pp. 309-312, 1991.
\bibitem[9]{panchapagesan16} Panchapagesan, S., Sun, M., Khare, A., Matsoukas, S., Mandal, A.,
Hoffmeister, B., and Vitaladevuni, S., ``Multi-task learning and weighted cross-entropy for dnn-based keyword spotting'', in \textit{Proceedings of Annual Conference of the International Speech Communication Association (Interspeech)}, 2016.
\bibitem[10]{chen14} Chen, G., Parada, C. and Heigold, G., ``Small-footprint keyword spotting using deep neural networks'', in \textit{IEEE International Conference on Acoustics, Speech and Signal Processing (ICASSP)},  pp. 4087-4091, 2014.
\bibitem[11]{nakkiran15} Nakkiran, P., Alvarez, R., Prabhavalkar, R. and Parada, C., ``Compressing deep neural networks using a rank-constrained topology", in \textit{Proceedings of Annual Conference of the International Speech Communication Association (Interspeech)}, 2015.
\bibitem[12]{sainath15} Sainath, T. and Parada, C., ``Convolutional neural networks for small-footprint keyword spotting", in \textit{Proceedings of Annual Conference of the International Speech Communication Association (Interspeech)}, 2015.
\bibitem[13]{tucker16} Tucker, G., Wu, M., Sun, M., Panchapagesan, S., Fu, G. and Vitaladevuni, S., ``Model
compression applied to small-footprint keyword spotting'', in \textit{Proceedings of Annual Conference of the International Speech Communication Association (Interspeech)}, 2016.
\bibitem[14]{fernndez07} Fernndez, S., Graves, A. and Schmidhuber, J., ``An application of recurrent neural networks to discriminative keyword spotting'', in \textit{Artificial Neural Networks-ICANN}, pp. 220-229, 2007.
\bibitem[15]{wollmer13} Wollmer, M., Schuller, B. and Rigoll, G., ``Keyword spotting exploiting long short-term memory'', in \textit{Speech Communication}, 55(2), pp.252-265, 2013.
\bibitem[16]{baljekar14} Baljekar, P., Lehman, J.F., and Singh, R., ``Online word-spotting in continuous speech with recurrent neural networks'', in \textit{IEEE Spoken Language Technology Workshop (SLT)}, 2014.
\bibitem[17]{sundar15} Sundar, H., Lehman, J.F. and Singh, R., 2015. ``Keyword spotting in multi-player voice driven games for children'', \textit{Proceedings of Annual Conference of the International Speech Communication Association (Interspeech)}, 2015.
\bibitem[18]{scherer10} Scherer, D., Muller, A. and Behnke, S., ``Evaluation of pooling operations in convolutional architectures for object recognition'', in \textit{Proceedings of International Conference on Artificial Neural Networks}, pp. 92-101, 2010.
\bibitem[19]{hochreiter97} Hochreiter, S. and Schmidhuber, J., ``Long Short-Term Memory'', in \textit{Neural Computation}, vol. 9, no. 8, pp. 1735-1780, 1997.
\bibitem[20]{gers02} Gers, F.A., Schraudolph, N.N. and Schmidhuber, J., ``Learning precise timing with LSTM recurrent networks'', in \textit{Journal of machine learning research}, vol. 3, pp. 115-143, 2002.
\bibitem[21]{sak14} Sak, H., Senior, A.W. and Beaufays, F., ``Long short-term memory recurrent neural network architectures for large scale acoustic modeling'', in \textit{Proceedings of Annual Conference of the International Speech Communication Association (Interspeech)}, 2014.
\bibitem[22]{bishop06} Bishop, C., ``Pattern Recognition and Machine Learning'', \textit{Springer}, 2006.
\bibitem[23]{graves05} Graves, A., Fernandez, S., Gomez, F. and Schmidhuber, J., ``Connectionist Temporal
Classification: Labelling Unsegmented Sequence Data with Recurrent Neural Networks'', in \textit{Proceedings of the International Conference on Machine Learning (ICML)}, 2006.
\bibitem[24]{ng15} Yue-Hei Ng, J., Hausknecht, M., Vijayanarasimhan, S., Vinyals, O., Monga, R. and Toderici, G., ``Beyond short snippets: Deep networks for video classification'', in \textit{Proceedings of the IEEE Conference on Computer Vision and Pattern Recognition (CVPR)}, 2015.
\bibitem[25]{xu15} Xu, Z., Li, S. and Deng, W., ``Learning temporal features using LSTM-CNN architecture for face anti-spoofing'', in \textit{Proceedings of IAPR Asian Conference on Pattern Recognition (ACPR)}, 2015, 
\bibitem[26]{strom15} Strom, N., ``Scalable Distributed DNN Training Using Commodity GPU Cloud Computing'', in \textit{Proceedings of Annual Conference of the International Speech Communication Association (Interspeech)}, 2015.
\end{thebibliography}

\end{document}